\documentclass[runningheads]{llncs}
\usepackage[T1]{fontenc}
\usepackage{graphicx, amsmath, multirow, booktabs, xcolor, float, subcaption}

\usepackage{threeparttable}

\begin{document}
\title{Dance Style Recognition Using Laban Movement Analysis}

\titlerunning{Dance Style Recognition using LMA}
%
\author{Muhammad Turab\inst{1} \and
Philippe Colantoni\inst{1}\orcidID{0000-0003-0002-4435} \and Damien Muselet\inst{1} \orcidID{0000-0001-7803-1171} \and Alain Tr\'{e}meau\inst{1}\orcidID{0000-0003-2826-7519}}
\authorrunning{M. Turab et al.}
%
\institute{
Laboratoire Hubert Curien - UMR 5516, Saint-Etienne, France \\
\email{muhammad.turab.muslim.bajeer@etu.univ-st-etienne.fr} \\
\email{\{philippe.colantoni, damien.muselet, alain.tremeau\}@univ-st-etienne.fr}
}
\maketitle              
\begin{abstract}
The growing interest in automated movement analysis has presented new challenges in recognition of complex human activities including dance. This study focuses on dance style recognition using features extracted using Laban Movement Analysis. Previous studies for dance style recognition often focus on cross-frame movement analysis, which limits the ability to capture temporal context and dynamic transitions between movements. This gap highlights the need for a method that can add temporal context to LMA features. For this, we introduce a novel pipeline which combines 3D pose estimation, 3D human mesh reconstruction, and floor aware body modeling to effectively extract LMA features. To address the temporal limitation, we propose a sliding window approach that captures movement evolution across time in features. These features are then used to train various machine learning methods for classification, and their explainability explainable AI methods to evaluate the contribution of each feature to classification performance. Our proposed method achieves a highest classification accuracy of 99.18\% which shows that the addition of temporal context significantly improves dance style recognition performance. 
\end{abstract}
\section{Introduction}
\label{sec:introduction}
Dance plays a vital role in human societies across time and cultures with different forms of artistic expression conveyed through different dance genres, dance styles and dance gestures. These diverse styles reflect not only cultural identity, but also emotions, stories, and social connections. Over the years, dance styles have evolved greatly reflecting cultural, social and artistic changes. Due to its rapid evolution, identification and recognition of various dance styles becomes increasingly challenging, often requiring the expert knowledge to differentiate them through visual analysis. Consequently, there is a growing need for robust analytical tools that can assist in identifying and classifying dance styles without relying solely on expert knowledge. One of the main framework used for movement analysis is Laban Movement Analysis (LMA) \cite{von1950mastery} which is a widely used framework for analyzing and interpreting human body movement. It consists of four main components: Body, Effort, Shape, and Space. These components cover the structural, spatial, and dynamic aspects of motion. The Body and Space components describe how the body moves either within itself or in relation to the surrounding 3D space. The Shape component focuses on how the body's form changes during movement, while the Effort component deals with the qualitative elements such as energy, dynamics, and intent. Although LMA is primarily qualitative, some research has aimed to develop feature descriptors that bring a level of quantification to these aspects. Still, challenges remain—especially in capturing the time-based progression of movements and providing clear, explainable and interpretable descriptors. Many current approaches struggle to reflect the full dynamic range of movement without considering temporal information. To address these limitations, we propose new feature descriptors that include temporal dynamics to better reflect the expressive qualities of LMA in both model predictions and individual performance. Our main contributions include:
\begin{itemize} 
\item Adding temporal dynamics to existing feature descriptors to better capture short-term movement patterns and trends. 
\item Introducing new descriptors to the four LMA components to represent movements more effectively. 
\item Evaluation of relevance and robustness of a set of feature descriptors from AIST++ dataset using machine learning models with their interpretation and explanation. 
\end{itemize}
\section{Related Work}
\label{sec:related_work}
Dance style recognition is a complex task that combines human motion analysis \cite{dos2013laban}, action classification \cite{ramezanpanah2020human}, and expressive movement understanding \cite{ajili2019expressive}. Expressive emotions in dance performances have been extensively studied in \cite{ajili2019expressive,ramezanpanah2020human,aristidou2014feature,aristidou2015emotion} using various methods, including LMA and deep learning. Recently, studies have also focused on dance style, genre, and gesture recognition. In \cite{sutopo2023dance}, LMA features were used for the recognition of dance gestures. \cite{dewan2018spatio} used Laban based spatio-temporal features for the classification of Indian dance style. \cite{guo2022ai} performed large-scale human motion classification for automatic movement annotation. \cite{baker2024computational} used various computational kinematic methods along with 3D pose estimation to classify hip hop dance movements. For automatic feature extraction and better results, Deep learning (DL) has been used. \cite{wang2020dance} applied a hybrid deep learning method based on Convolutional Neural Networks (CNN) and Long Short-Term Memory (LSTM) to effectively identify emotions. For body structure, spatial orientation, and force effect LMA-based feature descriptors were used to add emotional changes during movement. Their approach successfully recognizes emotions in dance movements with high accuracy.
\section{Methodology} 
\label{sec:methodology}
\begin{figure}[h!]
    \centering
    \includegraphics[width=\textwidth]{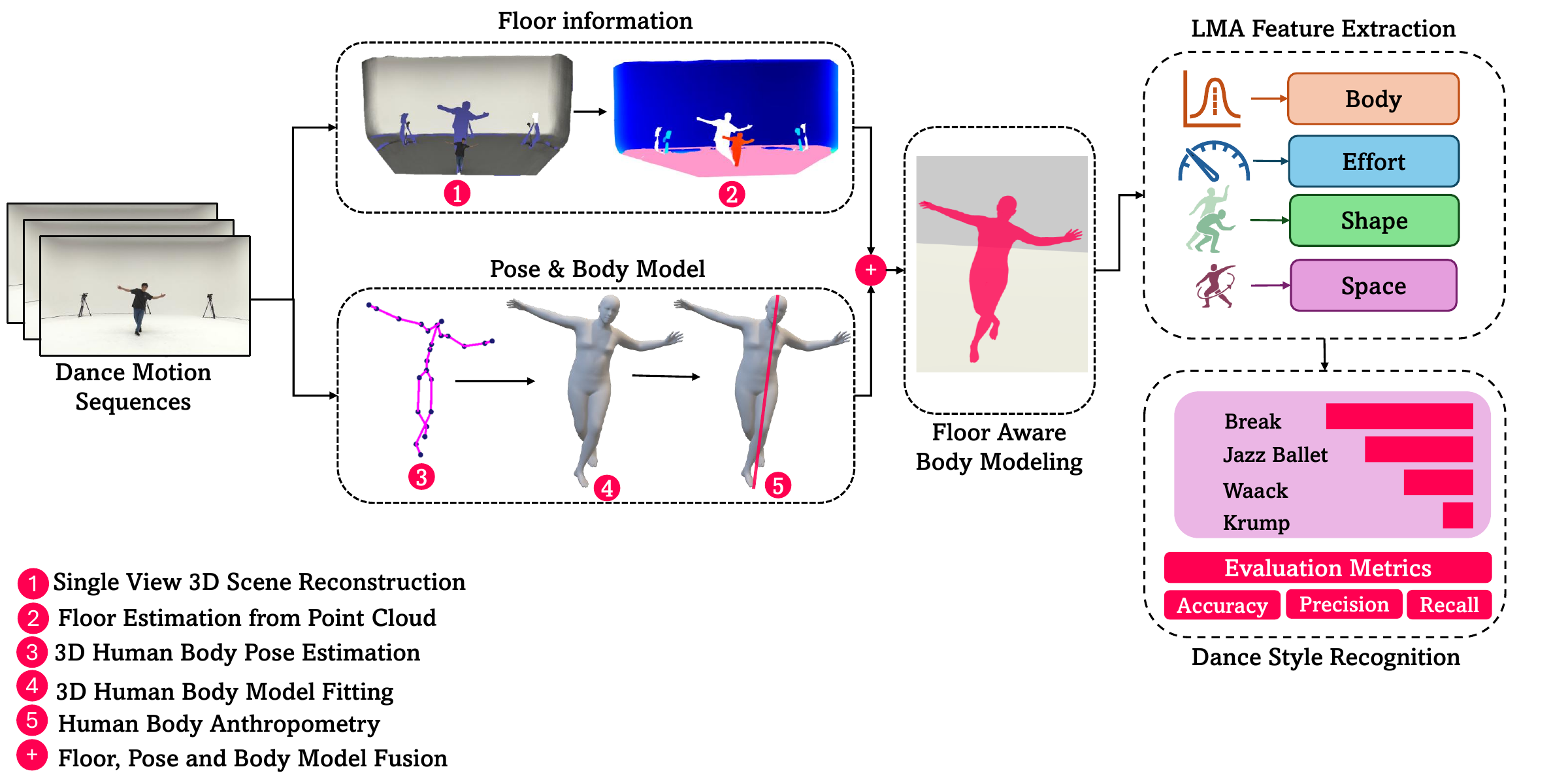}
    \caption{Overview of the proposed method for dance style recognition.}
    \label{fig:workflow}
\end{figure}

An overview of the proposed method is presented in Fig. \ref{fig:workflow}, which includes four key stages: 1) data preprocessing, 2) Floor aware body modeling, 3) LMA feature extraction, and 4) dance style recognition.
\vspace{-8mm}
\subsection{Dataset}
\label{subsec:data_processing}
The AIST Dance Video Database \cite{aist-dance-db} is a large scale collection of street dance videos. This database includes 10 street dance genres: Break, Pop, Lock, Waack, Middle Hip-Hop, LA-style Hip-Hop, House, Krump, Street Jazz, and Ballet Jazz. In this work, we limited our tests and experiments to dance videos in which a single dancer performs ("Basic" and "Advanced" dance), and we used only one single view, the frontal camera, as the source of information. The total number of dance videos considered in our experimental evaluation is 600. The resolution of the videos is 1920 × 1080 pixels with 60 frame rate.
\subsection{Floor Aware 3D Body Modeling}
\subsubsection{3D Human Body Pose Estimation}
To obtain the joints keypoint data from dance videos, we use Neural Localizer Fields (NLF) \cite{sarandi2024neural} for 3D body pose estimation. We compared NLF with other state-of-the-art (SOTA) methods including Mediapipe \cite{lugaresi2019mediapipe}, OpenPose \cite{cao2019openpose}, HMR2.0 \cite{goel2023}, ScoreHMR \cite{stathopoulos2024}, TokenHMR \cite{dwivedi2024}, OSX \cite{lin2023one}, SMPLer-X \cite{cai2024smpler} and choose NLF for several reasons including: 1) It is an advanced model trained on massive data to estimate 2D/3D points from single image, 2) it shows strong performance under occlusions and partial visibility, which frequently occur in dance videos when performers move close to the camera or self-occlude. 3) It provides joint coordinates in absolute camera space, effectively handling back-and-forth movements and shifting viewpoints and most of all, 4) for its ability to assess the complex poses that are associated with many types of dance.
\subsubsection{Floor Estimation from Point Cloud Data}
Many studies \cite{baker2024computational}\cite{ajili2017robust} assume the floor level by simply taking the minimum value of the ankle joint positions, particularly when computing features such as body height. However, this assumption often leads to inaccurate estimations. To enable more reliable analysis, we explicitly estimate the floor surface rather than assuming it. To achieve this, we first need to reconstruct the 3D scene from a single-view input. After testing several 3D reconstruction techniques: DUSt3R \cite{dust3r_cvpr24}, MASt3R \cite{mast3r_arxiv24}, VGGT \cite{wang2025vggt} we choose MoGe \cite{wang2024moge} for its quality of ground surface reconstruction. We then project the resulting point cloud onto the XZ-plane and apply quantile regression to fit a line that best represents the floor plane, effectively capturing any tilt or slope in the floor geometry.
\subsubsection{3D Human Body Model Fitting using SMPL}
Once we have the floor information, we fit a 3D human mesh using the SMPL model \cite{loper2023smpl}, based on the estimated 3D joint positions and shape parameters. We then apply pose-independent 3D anthropometry \cite{SMPL-Anthropometry} to compute body measurements, including the height of dancers. The dance videos considered in our work consist only of standing dancers.
\subsection{LMA Feature Extraction}
Laban Movement Analysis is a language and a structured framework to describe, and interpret human body movements. It categorizes human motion into four main components: Body, Effort, Shape and Space. \textbf{Body} component focuses on the physical structure and mechanics of the human body, including joint relationships, movement patterns, and inter-joint influences. To quantify joint connectivity, we compute Euclidean distances and inter-joint angles between key anatomical landmarks such as the hands, shoulders, pelvis, knees, and ankles. For modeling motion, we introduce a novel descriptor designed to detect the initiation of movement by a specific joint within a given time window, as shown in Equation \ref{eq:movement_initiation}.
\begin{equation}
\text{Initiation}(t) = \frac{P_j\bigl(t + w\bigr) - P_j\bigl(t\bigr)}{\Delta t} > \tau
\label{eq:movement_initiation}
\end{equation}

Here \(P_{j}(t)\) denotes the position of joint \(j\) at time \(t\), \(\Delta t\) is the time interval, \(w\) is the short time-window to add temporal dynamics, and \(\tau\) is a data-driven threshold calculated using standard-deviation of the entire sequence. \textbf{Effort} captures the dynamic qualities of movement, including the intention behind the motion, the energy applied, and its expressivity. We mainly focus on upper and lower body parts, such as hands, feet, head, and pelvis, as these joints are involved most in dance movements. Effort is broken down into four key factors: Space, Weight, Time, and Flow. Effort Space describes the attention of movement in space, and it can be direct where the movement is focused on single direction, and indirect where the movement is focused in multi-directions. For this we calculate the distance of each joint for a short time-window to the total distance covered by that joint using Equation \ref{eq:space_joint}.
\begin{equation}
\text{Space}_j(T) = \frac{\sum_{i=1}^{T} \left\| P_j(t_i) - P_j(t_i - w) \right\|}{\left\| P_j(T) - P_j(t_1) \right\|}
\label{eq:space_joint}
\end{equation}
Total space is calculated by multiplying the weights \(\alpha_j\)  of selected joints \(\mathcal{J}\) to their space factor using Equation \ref{eq:space_total}. Joint weights are used to give more importance to extremities, as defined in \cite{mmpose2020}.
\begin{equation}
\text{Space}(T) = \sum_{j \in \mathcal{J}} \alpha_j \, \text{Space}_j(T)
\label{eq:space_total}
\end{equation}
Effort Weight describes how powerful or strong a movement is, it can be strong or light. For this we calculate the kinetic energy of selected joints \(\mathcal{J}\) using Equation \ref{eq:effort_weight}. Here \(v_j(t_i)\) denotes the velocity of a joint $j$ at time $t_i$.
\begin{equation}
Weight(t) 
= \sum_{j \in \mathcal{J}} E_j(t_i) 
= \sum_{j \in \mathcal{J}} \frac{1}{2}\alpha_j v_j(t_i)^2
\label{eq:effort_weight}
\end{equation}
Effort Time captures a sense of urgency. It tells how quick or sustained a movement is executed. For this we calculate the acceleration of selected joints over a sliding time-window as shown in Equation \ref{eq:effort_time}.
\begin{equation}
\text{Time}(T) = \sum_{j \in \mathcal{J}} \alpha_j \, \text{Time}_j(T)
\label{eq:effort_time}
\end{equation}
where \(\text{Time}_j(T) = \frac{1}{T} \sum_{i=1}^{T} a_j(t_i)\), and $a_j$ denotes the acceleration of a joint $j$. \textbf{Space} component describes the relationship of movement with space. For this we calculate trajectory of whole sequence, curvature, and propose a new feature spatial dispersion that describes how dancer is using kinesphere or it's personal space. It is calculated as distance of upper body parts to torso, pelvis for lower body parts. total path covered, and total distance covered. \textbf{Shape} component describes how the body is changing shape during the movement. For this we calculate the volume of the body using ConvexHull algorithm implemented in Python SciPy package \cite{virtanen2020scipy}. 
Once we quantify all LMA components, we obtain a descriptor vector composed of 55 features which is used as an input to the classification models in the next section.
\subsection{Multi-class Dance Style Classification}
For dance style classification, we used machine learning methods including Support Vector Machines (SVM) \cite{cortes1995support} and Random Forests (RF) \cite{breiman2001random}, both of which have been used in previous studies on movements classification \cite{ajili2019expressive,wang2020dance,aristidou2015emotion}. We adopt 3-fold cross-validation to split the dataset into training, validation, and testing sets, helping to mitigate overfitting and reduce data dependency. For both SVM and RF, optimal hyperparameters are identified using Grid Search \cite{bergstra2012random}, with the best configuration selected based on validation accuracy.
\subsection{Model Explainability Using SHAP}
Machine learning models are often treated as black-box systems, where the reasoning behind their predictions is not easily interpretable. This lack of transparency makes it difficult to identify the features that influence their outputs. To enhance interpretability in the context of emotion classification for contemporary dance, we employ SHapley Additive exPlanations (SHAP) \cite{NIPS2017_7062}—a game theory-based approach that attributes importance values to each input feature for a given prediction. SHAP allows us to quantify the contribution of each feature to a model’s decision. For the SVM, we use the KernelExplainer, and TreeExplainer \cite{lundberg2020local2global} for RF.

\section{Results}
\begin{figure}[h!]
    \centering
    \includegraphics[width=0.7\textwidth]{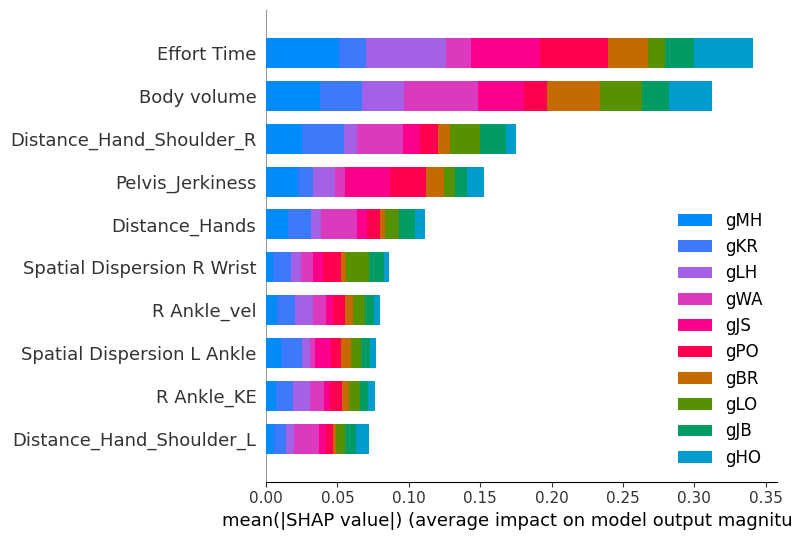}
    \caption{Impact and contribution of the top 10 features to the model predictions.}
    \label{fig:impact}
\end{figure}

\begin{figure}[h!]
    \centering
        \centering
        \includegraphics[width=0.7\textwidth]{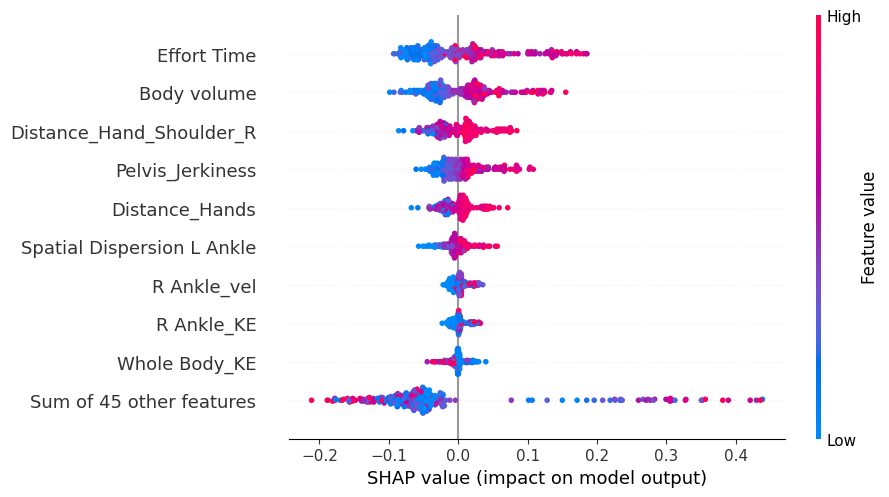}
        \caption{SHAP values illustrating feature contributions to the model’s predictions for the Middle Hip Hop dance style.}
        \label{fig:individual_influence_angry}
\end{figure}

\begin{figure}[h!]
        \centering
        \includegraphics[width=0.7\textwidth]{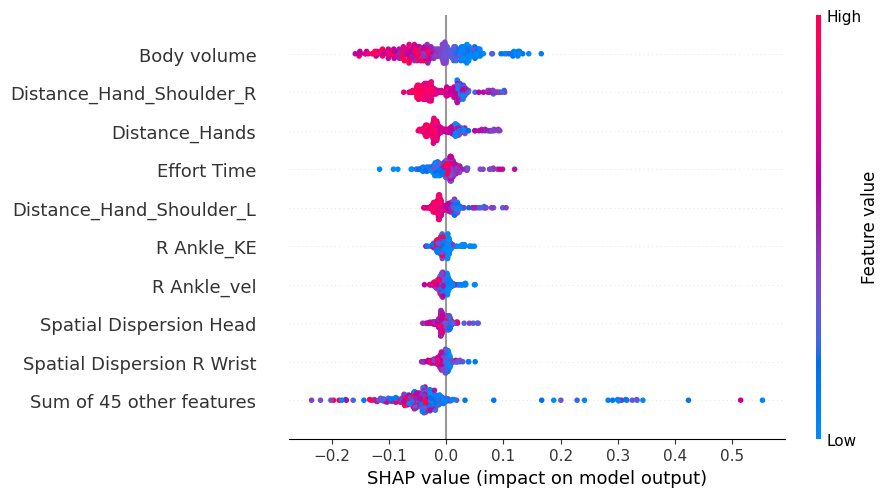}
        \caption{SHAP values illustrating feature contributions to the model’s predictions for the Waack dance style.}
    \label{fig:individual_influence_sad}
\end{figure}

In this section, we present the experimental results obtained using the proposed LMA feature descriptors. For evaluation, we used accuracy, precision, and recall metrics to assess the performance of our method. Next, we show the influence of each feature on the model predictions.
\vspace{-8mm}
\begin{table}[h!]
\centering
\setlength{\tabcolsep}{2pt}
\caption{Dance style classification results on the AIST++ dataset using a sliding window of 55 consecutive frames.}
\label{tab:classification_report}
\begin{tabular}{l@{\hskip 0.5cm}ccc@{\hskip 0.5cm}ccc}
\toprule
\multirow{2}{*}{\textbf{Style}} 
 & \multicolumn{3}{c}{\textbf{Random Forest}} 
 & \multicolumn{3}{c}{\textbf{SVM}} \\
\cmidrule(lr){2-4} \cmidrule(lr){5-7}
 & Prec. (\%) & Rec. (\%) & F1 (\%) & Prec. (\%) & Rec. (\%) & F1 (\%)\\
\midrule
Break (BR)  & 99.84 & 98.80 & 99.32 & 99.63 & 99.57 & 99.60\\
House (HO)  & 99.50 & 99.67 & 99.58 & 97.94 & 99.12 & 98.53 \\
Jazz Ballet (JB)  & 99.76 & 99.82 & 99.79 & 100.00 & 99.97 & 99.99 \\
Jazz Street (JS)  & 100.00 & 99.91 & 99.95 & 99.70 & 99.43 & 99.56 \\
Krump (KR)  & 99.88 & 99.27 & 99.58 & 99.27 & 98.49 & 98.87 \\
LA Hip Hop (LH)  & 99.68 & 99.91 & 99.79 & 99.26 & 99.17 & 99.22 \\
Lock (LO)  & 99.57 & 99.60 & 99.59 & 98.48 & 99.05 & 98.77 \\
Middle Hip Hop (MH)  & 98.34 & 99.66 & 99.00 & 98.66 & 99.15 & 98.90 \\
Pop (PO)  & 99.91 & 99.85 & 99.88 & 99.18 & 98.60 & 98.89 \\
Waack (WA)  & 99.85 & 99.88 & 99.87 & 99.32 & 99.26 & 99.29 \\
\midrule
\textbf{Average} & \textbf{99.63} & \textbf{99.64} & \textbf{99.64} & \textbf{99.14} & \textbf{99.18} & \textbf{99.16} \\
\bottomrule
\multicolumn{7}{l}{\small \textbf{Note: }Prec. = Precision, Rec. = Recall, and F1 = F1-score.} \\
\end{tabular}
\label{tab:all_emotions}
\end{table}
\vspace{-5mm}
Table \ref{tab:all_emotions} presents the results for each dance style. RF demonstrates consistent performance across all classes and metrics. Similarly, the SVM shows balanced performance, though lower than that of RF. The highest accuracy was achieved using a sliding window size of 55 consecutive frames, where RF achieved 99.68\% and SVM achieved 99.07\%. This window size likely captures a sufficient temporal context to model movement patterns effectively.
\vspace{-6mm}
\begin{figure}[h!]
    \centering
    \includegraphics[width=0.5\textwidth]{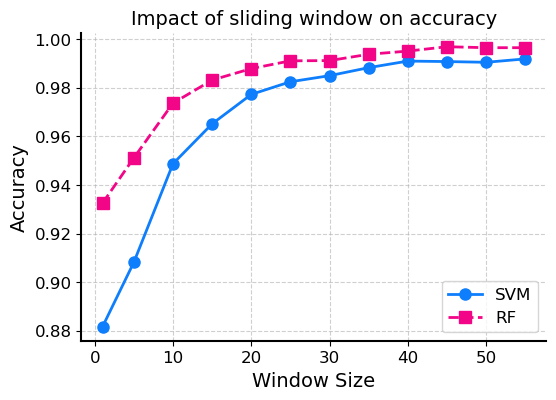}
    \caption{Impact of sliding window size on dance style classification accuracy.}
    \label{fig:sliding_window}
\end{figure}
\vspace{-5mm}
Fig. \ref{fig:sliding_window} shows the impact of window size on classification accuracy. It can be observed that the proposed sliding window approach significantly improves baseline performance between window sizes of 5 to 30 consecutive frames (which corresponds to a time interval of 3.5 $10^{-2}$ s to 0.5 s). After that, the improvement continues but is minimal. Although it is evident that a larger window improves accuracy, it is also important to note that it may suppress nuanced gesture details. Therefore, selecting an optimal window size is crucial. Although high classification accuracy indicates strong model performance, it is equally important to interpret how individual features influence the model's decisions. Fig. \ref{fig:impact} shows the most influential features for all styles. It is evident that features such as LMA effort time, body volume are most influential along with other low-level kinematic features, as they capture both the temporal dynamics and spatial characteristics of the dance styles. Fig. \ref{fig:individual_influence_angry} and \ref{fig:individual_influence_sad} show SHAP summary plots for the Middle Hip Hop and Waack classes and highlights the most influential features contributing to model predictions. For Middle Hip Hop Effort Time and Body Volume are dominant which indicates that temporal consistency and the spatial extent of the body has significant influence on  classification. Additional features such as Pelvis Jerkiness and Ankle Kinetics suggest that lower-body motion dynamics play a key role in distinguishing hip hop movements. In contrast for the Waack style, the most impactful features shift toward upper-body posture and extension. This reflects Waacking’s stylistic emphasis on expansive arm movements and structured upper body control. Notably, Effort Time has lower influence in this style, suggesting a stronger reliance on body shape and structure over motion.
\vspace{-5mm}
\subsection{Gesture Analysis and Style Comparison based on Kinematics}
\begin{figure}[h!]
    \centering
    \begin{subfigure}[b]{0.49\textwidth}
        \centering
        \includegraphics[width=\textwidth]{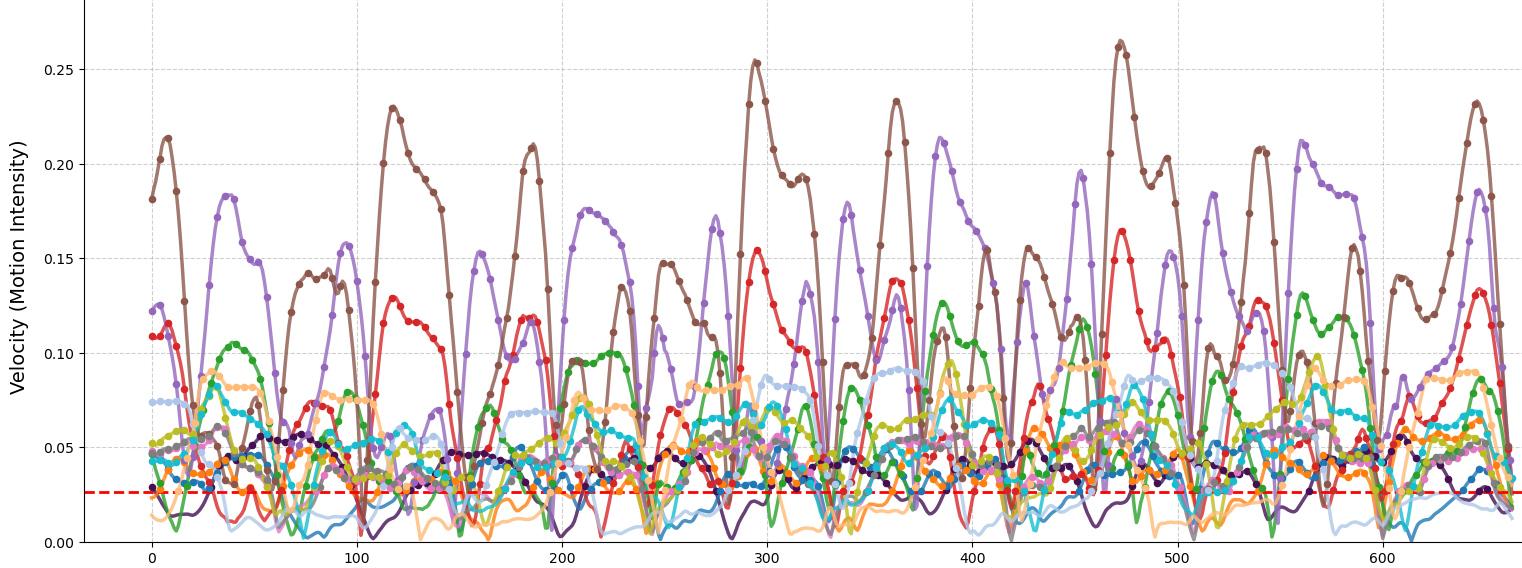}
        \caption{Break}
    \end{subfigure}
    \begin{subfigure}[b]{0.49\textwidth}
        \centering
        \includegraphics[width=\textwidth]{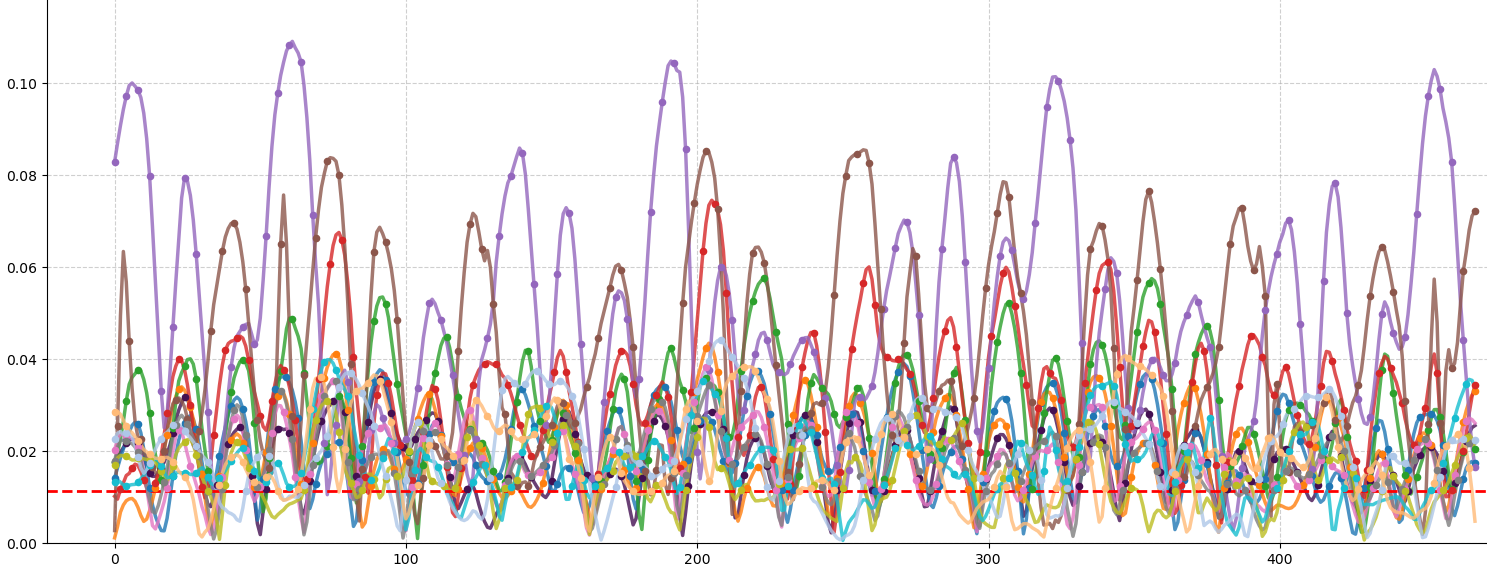}
        \caption{House}
    \end{subfigure}
     \begin{subfigure}[b]{0.49\textwidth}
        \centering
        \includegraphics[width=\textwidth]{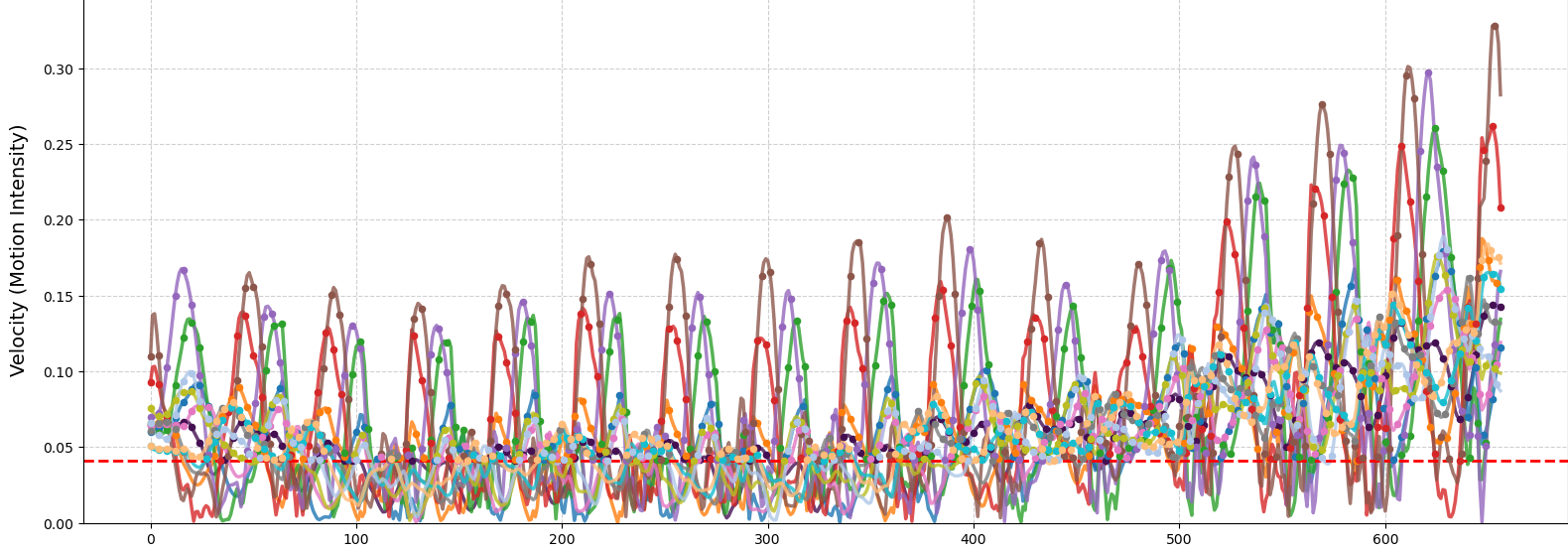}
        \caption{Jazz Ballet}
    \end{subfigure}
     \begin{subfigure}[b]{0.49\textwidth}
        \centering
        \includegraphics[width=\textwidth]{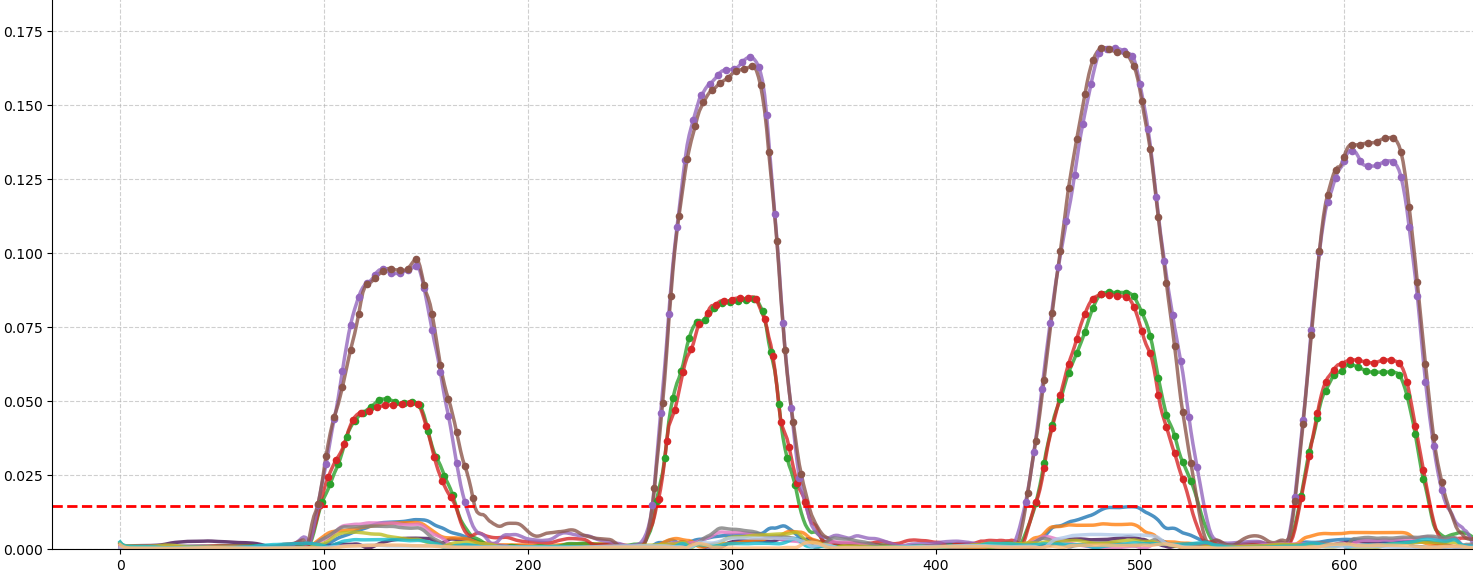}
        \caption{Jazz Street}
    \end{subfigure}
    \begin{subfigure}[b]{0.49\textwidth}
        \centering
        \includegraphics[width=\textwidth]{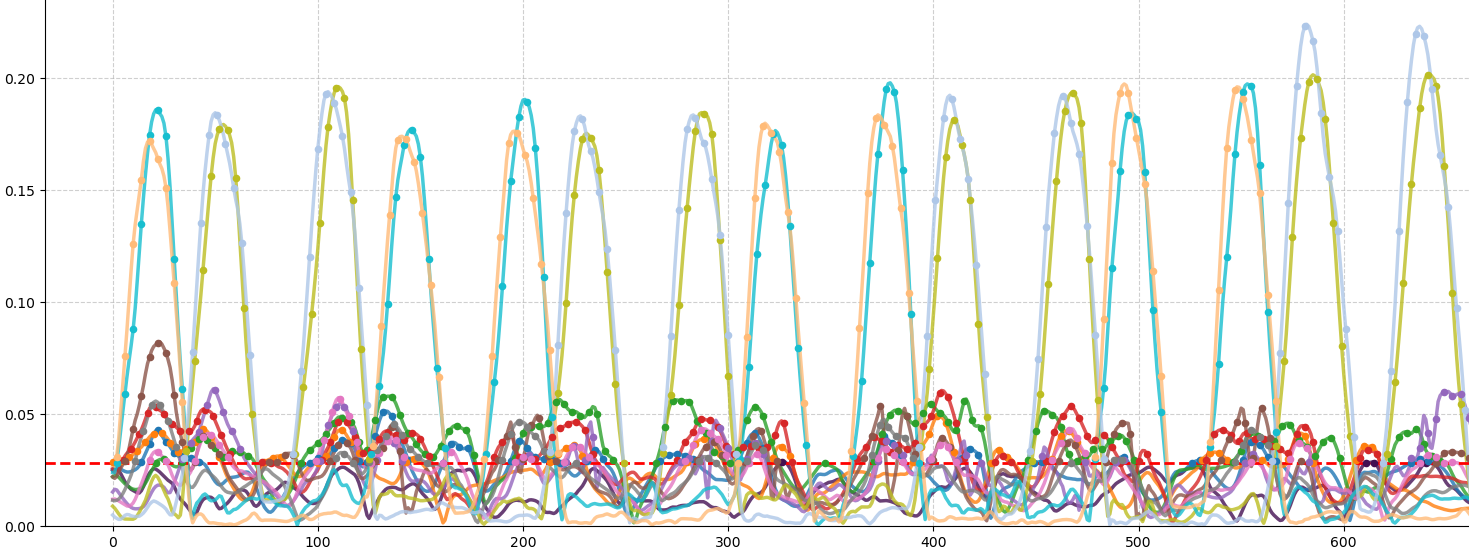}
        \caption{Krump}
    \end{subfigure}
    \begin{subfigure}[b]{0.49\textwidth}
        \centering
        \includegraphics[width=\textwidth]{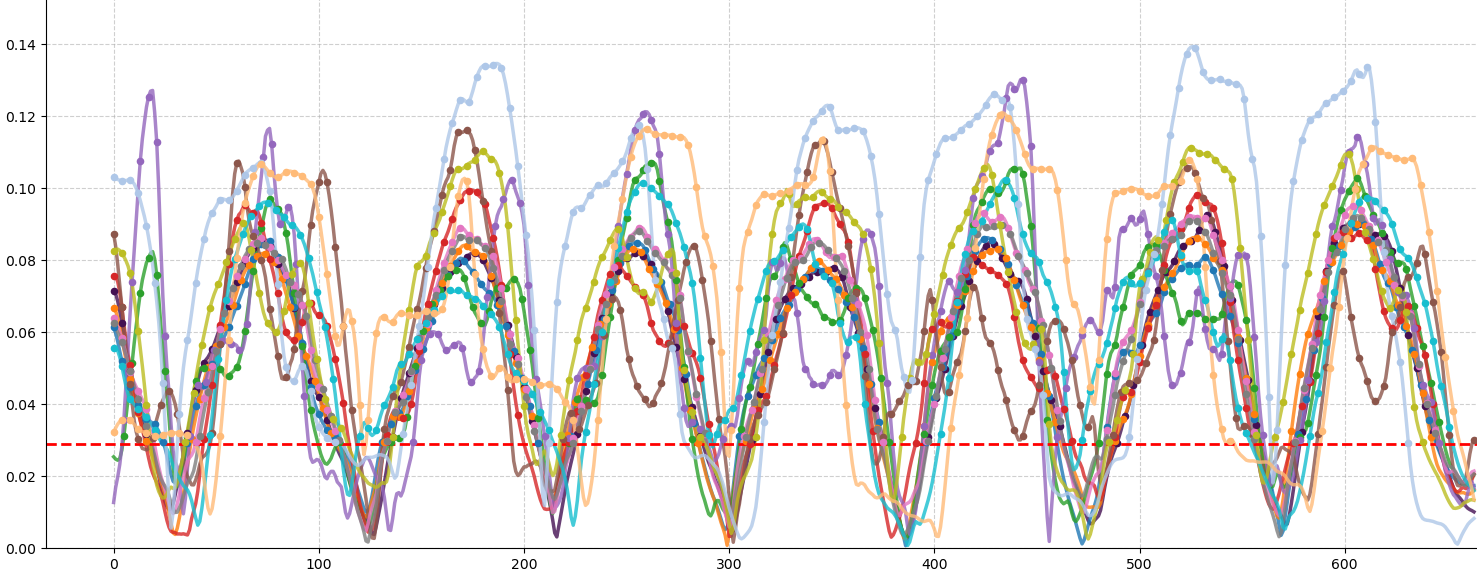}
        \caption{LA Hip Hop}
    \end{subfigure}
    \begin{subfigure}[b]{0.49\textwidth}
        \centering
        \includegraphics[width=\textwidth]{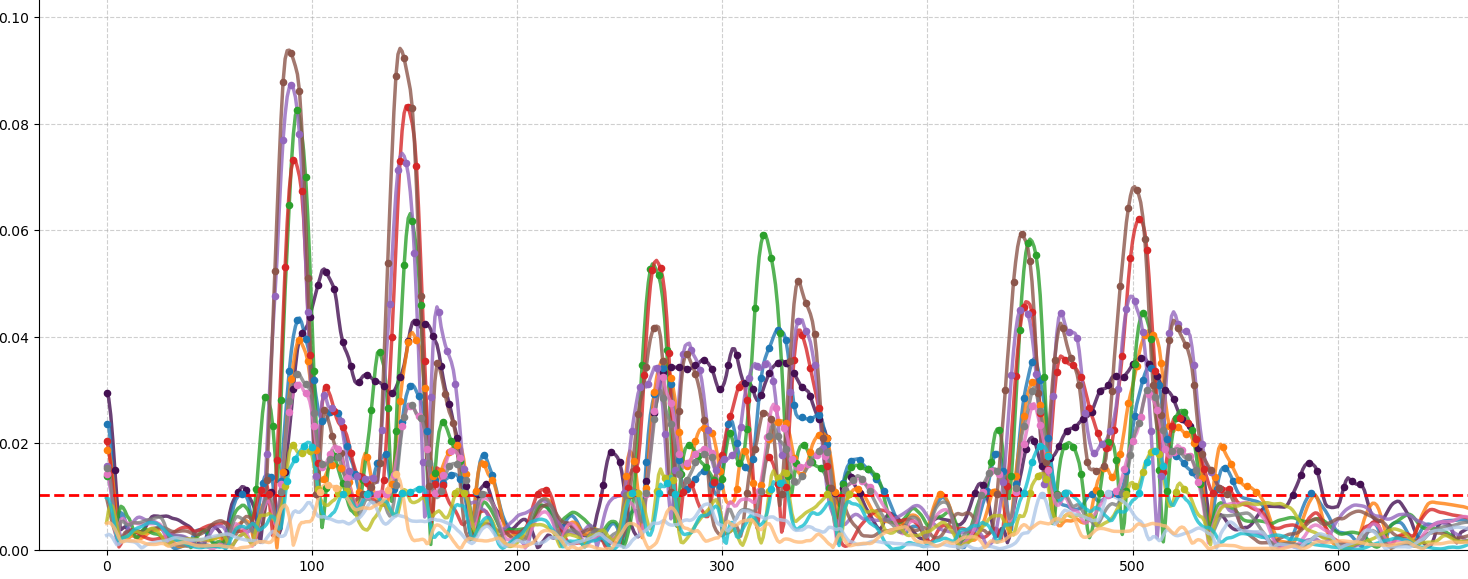}
        \caption{Lock}
    \end{subfigure}
    \begin{subfigure}[b]{0.49\textwidth}
        \centering
        \includegraphics[width=\textwidth]{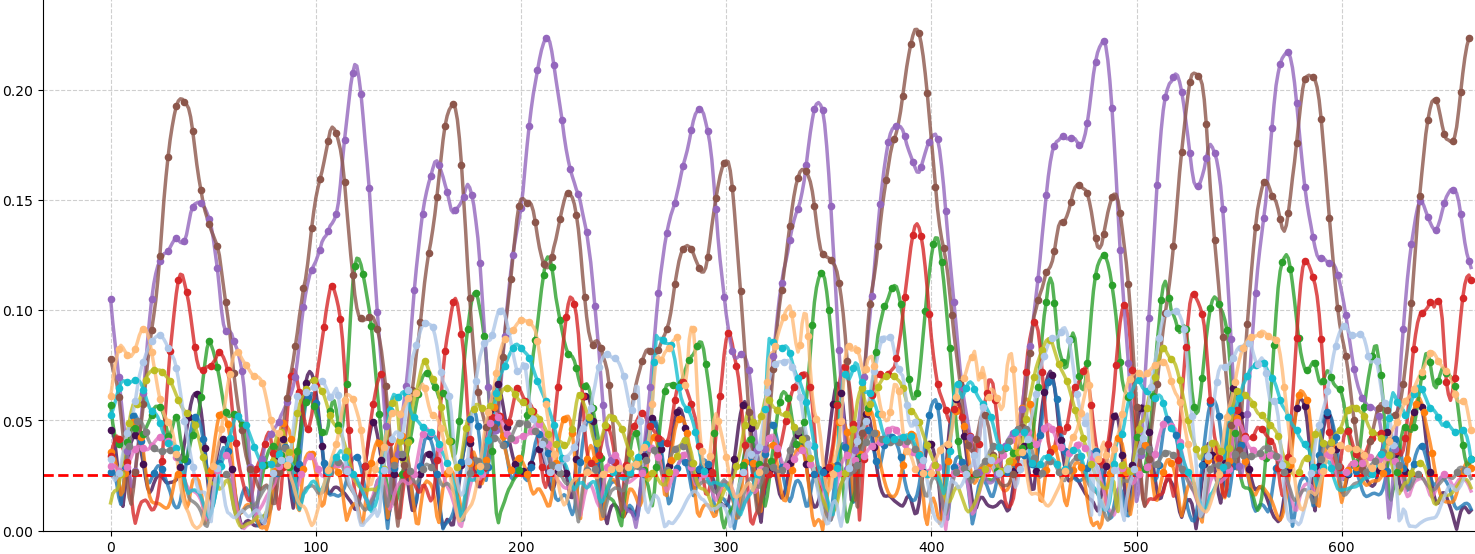}
        \caption{Middle Hip Hop}
    \end{subfigure}
    \begin{subfigure}[b]{0.49\textwidth}
        \centering
        \includegraphics[width=\textwidth]{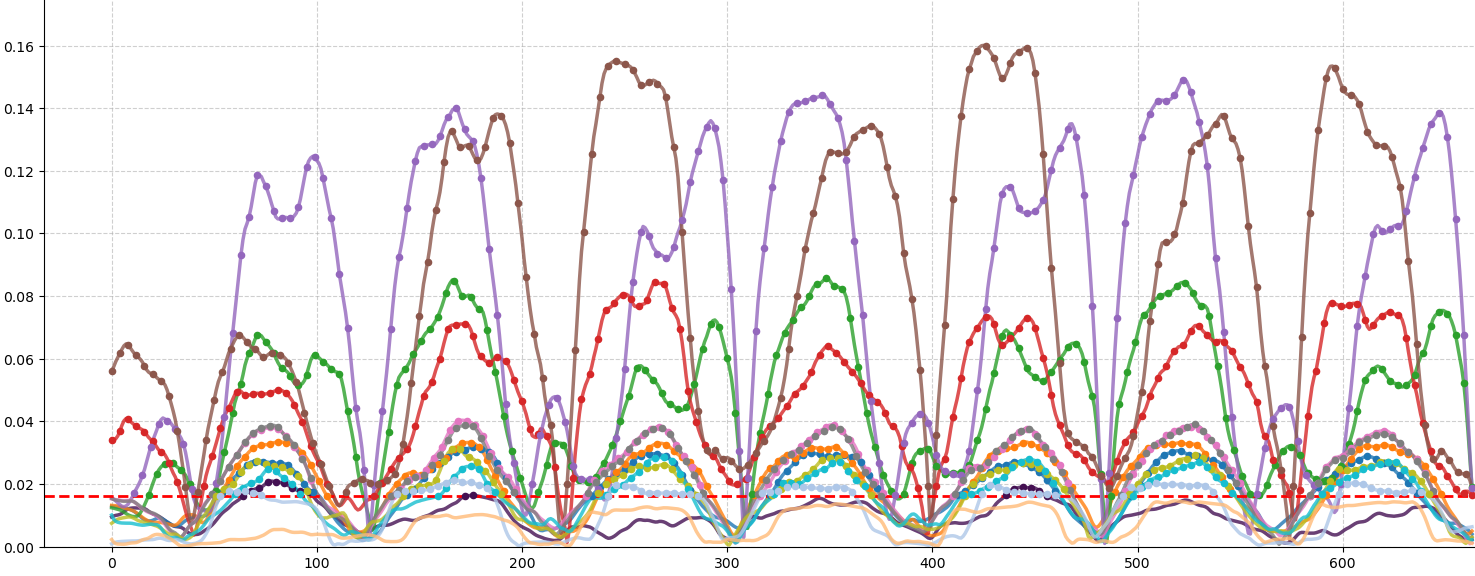}
        \caption{Pop}
    \end{subfigure}
    \begin{subfigure}[b]{0.49\textwidth}
        \centering
        \includegraphics[width=\textwidth]{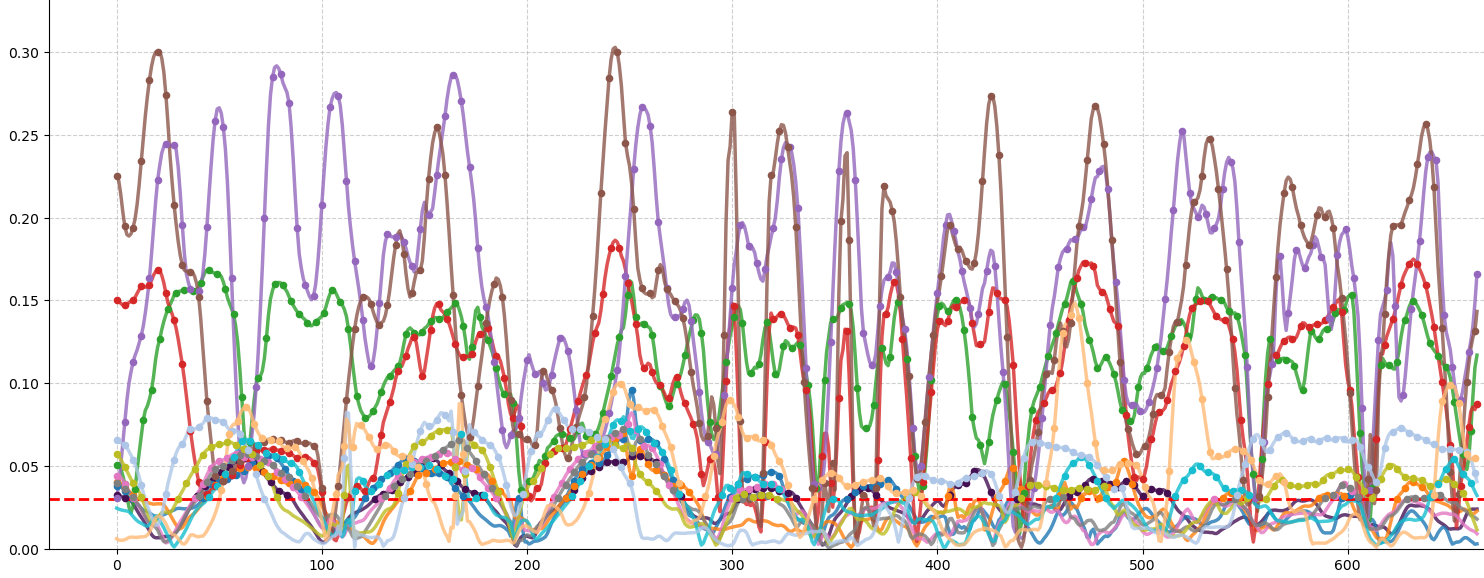}
        \caption{Waack}
    \end{subfigure}
    \caption{Long-term kinematic (velocity) evolution across ten dance styles, computed using a sliding window of 55 consecutive frames. The horizontal axis denotes the frame number, while the vertical axis indicates velocity.}
    \label{fig:kinematics}
\end{figure}

To better understand how movement evolves and how style-specific information is extracted using the proposed method, we present a gesture analysis of the ten dance styles based on kinematics, as shown in Fig. \ref{fig:kinematics}. Dance styles such as Ballet, Krump, LA Hip Hop, Lock, and Pop show repetitive and structured temporal patterns. For example, Krump and Pop show quick movements, while Lock shows clear gesture pause patterns. In contrast, Break, Waack, and Middle Hip Hop show random, varied, and less predictable temporal patterns. It is evident that using a larger sliding window size helps capture the style-specific representation better, which directly improves classification accuracy.
\vspace{-5mm}
\subsection{Feature Testing on Out-of-Distribution Dance Performances}
\begin{figure}[H]
        \centering
        \includegraphics[width=\textwidth]{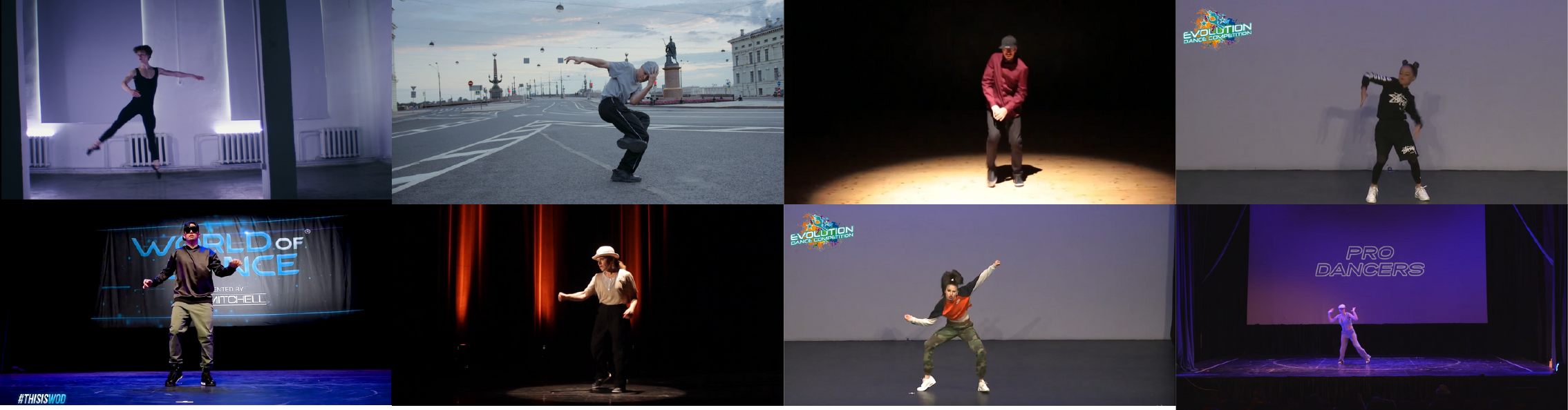}
        \caption{Examples of dance performance videos collected from online sources, used to evaluate the robustness of the extracted features.}
    \label{fig:robustness}
\end{figure}

To assess the robustness of the feature descriptors, we evaluated our classification method on a separate set of complex dance performances collected from online sources. These videos differ significantly from the training data of AIST++ on dance gestures, environment (illumination and background conditions), and performer variability; thus it provides a meaningful test of robustness, as shown in Fig. \ref{fig:robustness}. Our method performs well on some of the dance styles including lock, pop, ballet, LA hip hop, and waack, as these styles have unique dance gestures and movements (as seen in Fig. \ref{fig:kinematics}). For the remaining styles, such as mid hip-hop, break dance, house, jazz street, and krump, the performance decreases, as  these dances are characterized by more freedom of movement. This could be due to the nature of the collected dance videos from the internet, as the movements were unstructured, unlike the AIST++ dataset. Moreover, in AIST++
the performances were done by professionals. The highest accuracy was achieved on the lock style (81.89\% accuracy), and the lowest was on break dance (10.21\% accuracy). The strong performance on the lock style, despite being tested on out-of-distribution data, highlights that features capture and generalize well to structured dance patterns. These results show that like other ML methods, our solution is also dependent on the diversity and representativeness of the training dataset. Since AIST++ includes well-curated and structured dance sequences, the model struggles to generalize to some styles with more gesture, movement, and environmental variability that vary from the training distribution.
\vspace{-5mm}
\section{Conclusion}
The study introduces a new method that adds temporal dynamics to LMA feature descriptors to effectively capture the evolution of movement and long-term patterns, thereby improving the accuracy of dance style classification. In  future work, we will investigate solutions to enhance the robustness of our method against a diversity of performances for each dance style and a variety of dance video recording conditions.

\vspace{-3mm}
\subsubsection{Acknowledgments}
This study was carried out as part of the PREMIERE project "Performing Arts in a new Area", https://premiere-project.eu/, funded by HORIZON-CL2-2021-HERITAGE-000201-04 (grant number 101061303 - PREMIERE). 
\vspace{-3mm}
\bibliographystyle{splncs04}
\bibliography{main}
\end{document}